\documentclass{article}

\usepackage{microtype}
\usepackage{graphicx}
\usepackage{subfigure}
\usepackage{booktabs} 

\usepackage{times}
\usepackage{soul}
\usepackage{url}
\usepackage[hidelinks]{hyperref}
\usepackage[utf8]{inputenc}
\usepackage[small]{caption}
\usepackage{amsmath}
\usepackage{amsthm}
\usepackage{amssymb,amsfonts}
\usepackage{color}
\usepackage{multirow,multicol}
\usepackage{soul}
\usepackage{diagbox}
\makeatletter
\newcommand*\bigcdot{\mathpalette\bigcdot@{.5}}
\newcommand*\bigcdot@[2]{\mathbin{\vcenter{\hbox{\scalebox{#2}{$\m@th#1\bullet$}}}}}

\usepackage{hyperref}

\usepackage[accepted]{icml2020}

\icmltitlerunning{FL-AGCNS: Federated Learning Framework for Automatic Graph Convolutional Network Search}

\begin{document}

\twocolumn[
\icmltitle{FL-AGCNS: Federated Learning Framework for Automatic Graph Convolutional Network Search}



\icmlsetsymbol{equal}{*}

\begin{icmlauthorlist}
\icmlauthor{Chunnan Wang}{equal,goo}
\icmlauthor{Bozhou Chen}{goo}
\icmlauthor{Geng Li}{goo}
\icmlauthor{Hongzhi Wang}{equal,goo,ed}
\end{icmlauthorlist}

\icmlaffiliation{goo}{Harbin Institute of Technology, Harbin, China}
\icmlaffiliation{ed}{Peng Cheng Laboratory, Shenzhen, China}

\icmlcorrespondingauthor{Chunnan Wang}{WangChunnan@hit.edu.cn}
\icmlcorrespondingauthor{Bozhou Chen}{bozhouchen@hit.edu.cn}
\icmlcorrespondingauthor{Geng Li}{1173710214@stu.hit.edu.cn}
\icmlcorrespondingauthor{Hongzhi Wang}{wangzh@hit.edu.cn}

\icmlkeywords{Machine Learning, ICML}

\vskip 0.3in
]



\printAffiliationsAndNotice{}  

\begin{abstract}
Recently, some \underline{N}eural \underline{A}rchitecture \underline{S}earch (NAS) techniques are proposed for the automatic design of \underline{G}raph \underline{C}onvolutional \underline{N}etwork (GCN) architectures. They bring great convenience to the use of GCN, but could hardly apply to the \underline{F}ederated \underline{L}earning (FL) scenarios with distributed and private datasets, which limit their applications. Moreover, they need to train many candidate GCN models from scratch, which is inefficient for FL. To address these challenges, we propose FL-AGCNS, an efficient GCN NAS algorithm suitable for FL scenarios. FL-AGCNS designs a federated evolutionary optimization strategy to enable distributed agents to cooperatively design powerful GCN models while keeping personal information on local devices. Besides, it applies the GCN SuperNet and a weight sharing strategy to speed up the evaluation of GCN models. Experimental results show that FL-AGCNS can find better GCN models in short time under the FL framework, surpassing the state-of-the-arts NAS methods and GCN models.
\end{abstract}

\section{Introduction}

\underline{G}raph \underline{C}onvolutional \underline{N}etwork (GCN) is a powerful deep learning approach for graph-structured data. It can learn high-level node representation from node features and linkage patterns, thus effectively deal with graph-based tasks, such as node classification~\cite{p1}, traffic forecasting~\cite{p3} and online recommendation~\cite{p4}. Despite great success of GNNs, the design of GCN architecture requires both heavy manual work and domain knowledge~\cite{p5}, which is very laborious.

In order to reduce the development cost of GCNs, recently researches develop some \underline{GCN} \underline{N}eural \underline{A}rchitecture \underline{S}earch (GCN NAS) algorithms, including GraphNAS~\cite{p5} and SNAG~\cite{p6}, to automatically discover good GCN architectures. GraphNAS and SNAG apply a recurrent neural network controller to sample candidate architectures from the search space, and train controller with policy gradient to maximize the expected validation accuracy of the generated architectures. 
These methods greatly reduce the labour of human experts and can find better GCN  architectures than the human-invented ones. However, existing approaches are hardly applicable to the \underline{F}ederated \underline{L}earning (FL) scenarios~\cite{p18} with distributed and private graph datasets due to lack of information exchange scheme and efficient evaluation method, which reduces their practicality.

Specifically, the existing GCN NAS techniques are centralized learning approaches, i.e., they require the graph dataset to be aggregated on a single machine or in a datacenter. However, in many practical scenarios, the graph data is distributed across multiple clients without data sharing. For example, the traffic graph with flow data of a country is distributed across different cities, and the social graph with users' purchase history stored by a multinational corporation is scattered across different datacenters. These graph data fail to be aggregated due to huge transmission cost or information privacy, but a large amount of graph information and labels are still needed for getting a robust GCN model. This requires GCN NAS techniques to be able to cope with FL scenarios, i.e., design effective information exchange strategy and thus learn the optimal GCN architecture in a distributed and privacy-preserving manner. However, existing solutions fail to do so, which is not practical enough.

In addition, we notice that the GCN evaluation methods applied in the existing GCN NAS solutions are inefficient for the FL scenario. They need to train numerous architecture candidates from scratch and for a large number of epochs. In the FL framework, such methods bring huge amount of communications, since architectures should be trained jointly by multiple clients through information exchange for each epoch, which is very inefficient.


In this paper, we address the above challenges, and propose FL-AGCNS, an efficient GCN NAS algorithm, that enables distributed agents to cooperatively design powerful GCN models while keeping personal information on local devices.

Specifically, FL-AGCNS designs a federated evolutionary optimization strategy to fully consider the preferences of each client, and thus recommends GCN architectures that perform well in multiple datasets. Specifically, in the early evolution, we find that population individuals may not be applicable to some clients and thus show poor performance in the FL framework. To efficiently improve the overall performance of individuals, we execute evolutionary algorithms in clients to explore the characteristics of multiple datasets, and utilize the GCN architectures favored by each client to guide individuals to efficiently cover their shortages.

Besides, FL-AGCNS applies the GCN SuperNet, a weight sharing strategy, to speed up the model evaluation. It does not train different GCN architectures from scratch separately in the search stage, but to optimize parameters of GCN SuperNet, and efficiently evaluate various GCN architectures by sharing corresponding parameters in SuperNet. Such weight sharing strategy dramatically reduces the computational complexity of FL-AGCNS, making the search stage of FL-AGCNS efficient. The experimental results show that our optimization strategy performs well in the FL framework, outperforming the traditional evolutionary strategy which only considers the overall performance scores during the evolution. The code will be available on Github~\footnote{Our Github link will be available in the published version.}.

Our major contributions are concluded as follows:
\begin{itemize}
\vspace{-0.5cm}
\item Innovation: We are the first to enable the GCN NAS technique to run on the FL framework. The combination of GCN NAS and FL strengthens the practicality of the GCN NAS method 
and increases the flexibility of the FL framework.
\vspace{-0.35cm}
\item Effectiveness: We design a federated evolutionary optimization strategy to effectively search for high-quality GCN architectures under FL framework. 
\vspace{-0.35cm}
\item Efficiency: We propose to use GCN SuperNet to reduces the search cost of the GCN NAS method, promoting the search efficiency.
\vspace{-0.0cm}
\end{itemize}


\section{Prerequisite}\label{section:2}
We firstly introduce the existing GCNs and FL techniques, then give related concepts of FL based GCN NAS.

\subsection{Graph Convolutional Network}\label{section:2.1}

\underline{G}raph \underline{C}onvolutional \underline{N}etworks (GCNs) are a kind of neural networks which generalize the operation of convolution from grid data to graph data~\cite{p7}. 
The existing GCNs fall into two categories, spectral-based methods~\cite{p9,p10,p11} which define graph convolutions by introducing filters from the perspective of graph signal processing~\cite{p12}, and spatial-based methods~\cite{p13,p14,p15,p16} which define graph convolutions by information propagation. Two kinds of methods interpret graph convolutions from different angles, and propose many effective graph convolutional layers to extract node features from the one-hop neighborhood. In this paper, we flexibly use these existing graph convolutional layers to achieve the automatic design of appropriate GCN architectures.

\subsection{Federated Learning}\label{section:2.2}

\underline{F}ederated \underline{L}earning (FL) is a decentralized approach, which aims to train robust centralized model based on datasets that are distributed across multiple clients without sharing their data~\cite{p17,p18}. FL makes it possible to vigorously develop AI techniques in privacy-preserving era, and attracts great attention of scholars. Many effective FL schemes are proposed to train well-known machine learning models, such as deep neural networks~\cite{p19} and random forest~\cite{p20}, and achieve great effect in the real applications. However, there is still no FL work on automatic model design yet. In this paper, we fill this gap, enabling FL framework to automatically learn good GCN architectures.


Based on distribution characteristics of the data, FL algorithms can be divided into three categories, i.e., horizontal FL, vertical FL and federated transfer learning~\cite{p18}. Horizontal FL applies to scenarios that datasets share same feature space but different in samples~\cite{p21}, whereas the other two categories consider datasets with different feature spaces. In this paper, we focus on the cases that graph dataset is distributed to different regions in the form of densely connected subgraphs, nodes are different among clients, but the feature space of each node is same. Therefore, our study belongs to horizontal FL. Our study can be extended to others FL scenarios after minor adjustments, which will be studied in our future works.


\begin{figure*}[t]
    \centering
    \includegraphics[width=0.99\textwidth]{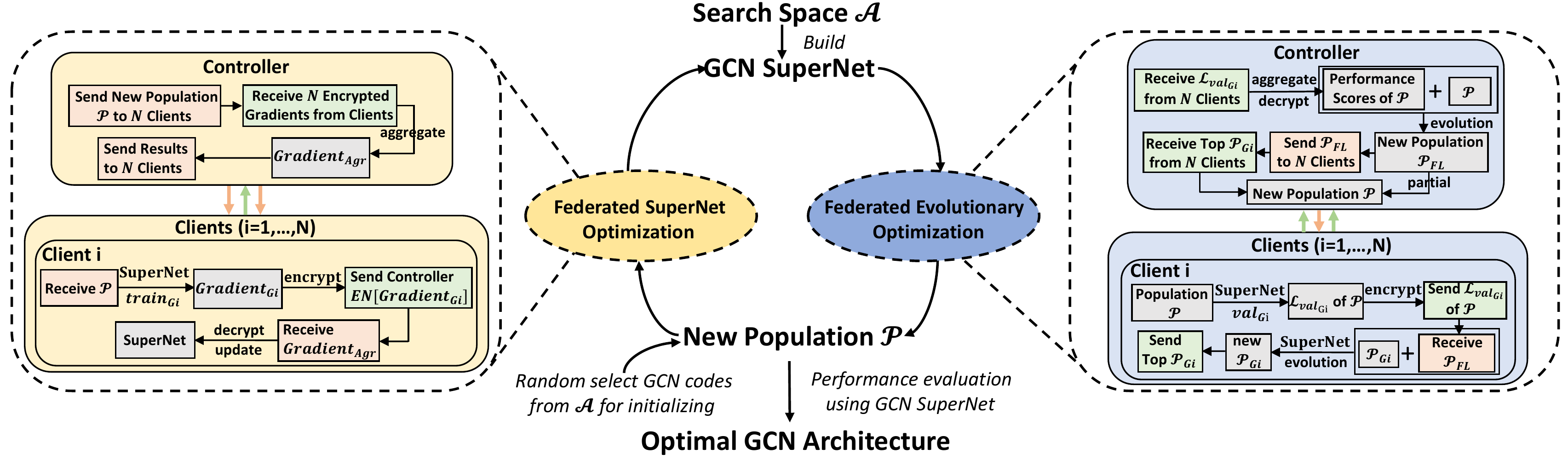}
	 \vspace{-0.35cm}
    \caption{Overall framework of FL-AGCNS. Subgraphs are scattered in different clients without sharing. Each client is responsible for providing its preferred architectures and local evaluation/gradient information, and the controller aims to guide clients to search for the optimal GCN architecture by analyzing or aggregating these messages.}
\label{fig1}
\vspace{-0.32cm}
\end{figure*}

\subsection{Related Definitions}\label{section:2.3}

We introduce graph's basic concepts, and describe the horizontal FL based GCN NAS problem studied in this paper.

\textbf{Definition 1: Graph and Graph Signal.} We use $G=(V_G,A_G,X_G)$ to denote a graph, where $V_G$ is the set of nodes, $A_G\in R^{N\times N}$ is the adjacency matrix of $G$, $X_G\in R^{N\times F}$ is the graph signal (feature matrix) of the graph $G$, $|V_G|=N$ denotes the number of vertices, and $F$ denotes the number of attribute features of each node. We use $train_{G}$, $val_{G}$ and $test_{G}$ to represent the training set, validation set and test set of graph $G$, respectively. 


\textbf{Definition 2: HFL based GCN NAS.} Given a search space for GCN architectures $\mathcal{A}$ and a set of subgraphs $G_i$  $(i=1,…,N)$ that are distributed and non-shared across $N$ clients, the \underline{H}orizontal \underline{FL} \underline{based} \underline{GCN} \underline{NAS} (HFL based GCN NAS) problem aims to find $\alpha^{\ast}\in \mathcal{A}$ that minimizes the overall validation loss of all subgraphs.
\vspace{-0.35cm}
\begin{equation}
\vspace{-0.5cm}
\small
\alpha^{\ast} = \min_{\alpha \in \mathcal{A}} \sum_{i=1}^{N} \frac{|val_{G_i}|}{\sum\limits_{j=1}^{N} |val_{G_j}|} \times \mathcal{L}_{val_{G_i}}(w^{\ast}(\alpha),\alpha)
\notag
\vspace{-0.3cm}
\end{equation}
\begin{equation}
\small
s.t.\ w^{\ast}(\alpha)=\mathop{argmin}\limits_{w} \sum_{i=1}^{N} \frac{|train_{G_i}|}{\sum\limits_{j=1}^{N} |train_{G_j}|}\times \mathcal{L}_{train_{G_i}}(w,\alpha)
\notag
\vspace{-0.35cm}
\end{equation}
where $\mathcal{L}_{val_{G}} (w,\alpha)$ and $\mathcal{L}_{train_{G}} (w,\alpha)$ represent the validation and training loss of $\alpha$ on graph $G$ under weights $w$ respectively. Different from the traditional GCN NAS, which only copes with a complete graph $G$. HFL-based GCN NAS breaks the limitation of ``Isolated Data Islands'', enabling the scattered graph data to jointly search for good GCN architectures without leaking private information.


\section{Our Method}\label{section:3}

In this paper, we design the FL-AGCNS algorithm to deal with the HFL based GCN NAS problem. We apply GCN SuperNet to achieve fast evaluation of GCN architectures (Section~\ref{section:3.2}), and utilize the federated evolutionary optimization strategy (Section~\ref{section:3.3}) to effectively improve the federated performance of the population. Following previous work~\cite{p24,p25}, we optimize SuperNet weights and population successively in each training step, and thus reduce expensive inner optimization of our bi-optimization problem.
Figure~\ref{fig1} gives the overall framework of FL-AGCNS, and the following are the detailed introduction of FL-AGCNS.

\begin{table*}[t]
\caption{Three stages that need to be gown through to construct a GCN architecture.}
\newcommand{\tabincell}[2]{\begin{tabular}{@{}#1@{}}#2\end{tabular}}
\centering
\resizebox{0.97\textwidth}{!}{
\smallskip\begin{tabular}{|m{2.3cm}|m{5.3cm}|m{3.2cm}|l|}
\hline
\textbf{Stage} & \textbf{Stage Function Description} & \textbf{Detail Contents} & \textbf{Operation Options} \\
\hline
\tabincell{l}{Stage1: Input\\ Transform Stage} & Produce low-dimensional representations of nodes. & Input Structure (IS) & \tabincell{l}{$\bigcdot$ $IS_{1}$: FullyConnection + Sigmoid (out size:64)\\ $\bigcdot$ $IS_{2}$: FullyConnection + Tanh (out size:64)\\ $\bigcdot$ $IS_{3}$: FullyConnection + Relu (out size:64)\\ $\bigcdot$ $IS_{4}$: FullyConnection + Softmax (out size:64)\\ $\bigcdot$ $IS_{5}$: FullyConnection + Identity (out size:64)} \\
\hline
\multirow{2}{*}{\tabincell{l}{\\ Stage2:  Feature \\Embedding Stage}} & Get high-level node features using GCN layers: $\{GCNL_{i} | i=1,…,L\}$. & GCN Type of $GCNL_{i}$ ($LType_{i}$) & \tabincell{l}{$\bigcdot$ $LType_{1}$: GATConv\ \ \ $\bigcdot$ $LType_{2}$: GINConv\ \ \ $\bigcdot$ $LType_{3}$: SAGEConv\\ $\bigcdot$ $LType_{4}$: GCNConv\ \ \ $\bigcdot$ $LType_{5}$: SGConv\ \ \ $\bigcdot$ $LType_{6}$: APPNP\\ $\bigcdot$ $LType_{7}$: AGNNConv\ \ \ $\bigcdot$ $LType_{8}$: ARMAConv\ \ \ $\bigcdot$ $LType_{9}$: FeaStConv\\ $\bigcdot$ $LType_{10}$: GENConv\ \ \ $\bigcdot$ $LType_{11}$: GMMConv\ \ \ $\bigcdot$ $LType_{12}$: GatedGraphConv} \\
\cline{3-4}
& \tabincell{l}{Note: $L$ is the number of GCN layers,\\ and detail contents of each GCN layer\\ $GCNL_{i}$ are described in the right part.} & \tabincell{l}{Index of the Preceding\\ GCN Layer of $GCNL_{i}$\\ ($LPreIndex_{i}$)} & \tabincell{l}{$LPreIndex_{i}$=$\{$None,0,…,i-1$\}$\\ Note: $LPreIndex_{i}$=None denotes $GCNL_{i}$,…,$GCNL_{N}$ are invalid, and\\ the output of Stage 2 is considered as $Mean(GCNL_{1},…,GCNL_{i-1})$.}\\
\hline
\tabincell{l}{Stage3: Output\\ Transform Stage} & Transform the output of the Stage2 (S2) into the expected prediction. & Output Structure (OS) & \tabincell{l}{$\bigcdot$ $OS_{1}$: FullyConnection + Sigmoid (out size:64)\\ $\bigcdot$ $OS_{2}$: FullyConnection + Tanh (out size:64)\\ $\bigcdot$ $OS_{3}$: FullyConnection + Relu (out size:64)\\ $\bigcdot$ $OS_{4}$: FullyConnection + Softmax (out size:64)\\ $\bigcdot$ $OS_{5}$: FullyConnection + Identity (out size:64)} \\
\hline
\end{tabular}
}
\label{table1}
\vspace{-0.4cm}
\end{table*}

\subsection{Code Representation of GCN in FL-AGCNS}\label{section:3.1}

The design of a GCN architecture contains the following 3 stages: (1) Input transform stage, which produces low-dimensional representations of nodes; (2) Feature embedding stage, which extracts high-level node features using multiple GCN layers; (3) Output transform stage, which transforms the output of the feature embedding stage to the final prediction. In FL-AGCNS we use $(2\times L+2)$ parameters ($L$ is the number of GCN layers), as is shown in Table~\ref{table1}, to describe the detailed structures of these stages, so as to obtain the code representation of the GCN architecture.

\begin{figure}[t]
    \centering
    \includegraphics[width=0.485\textwidth]{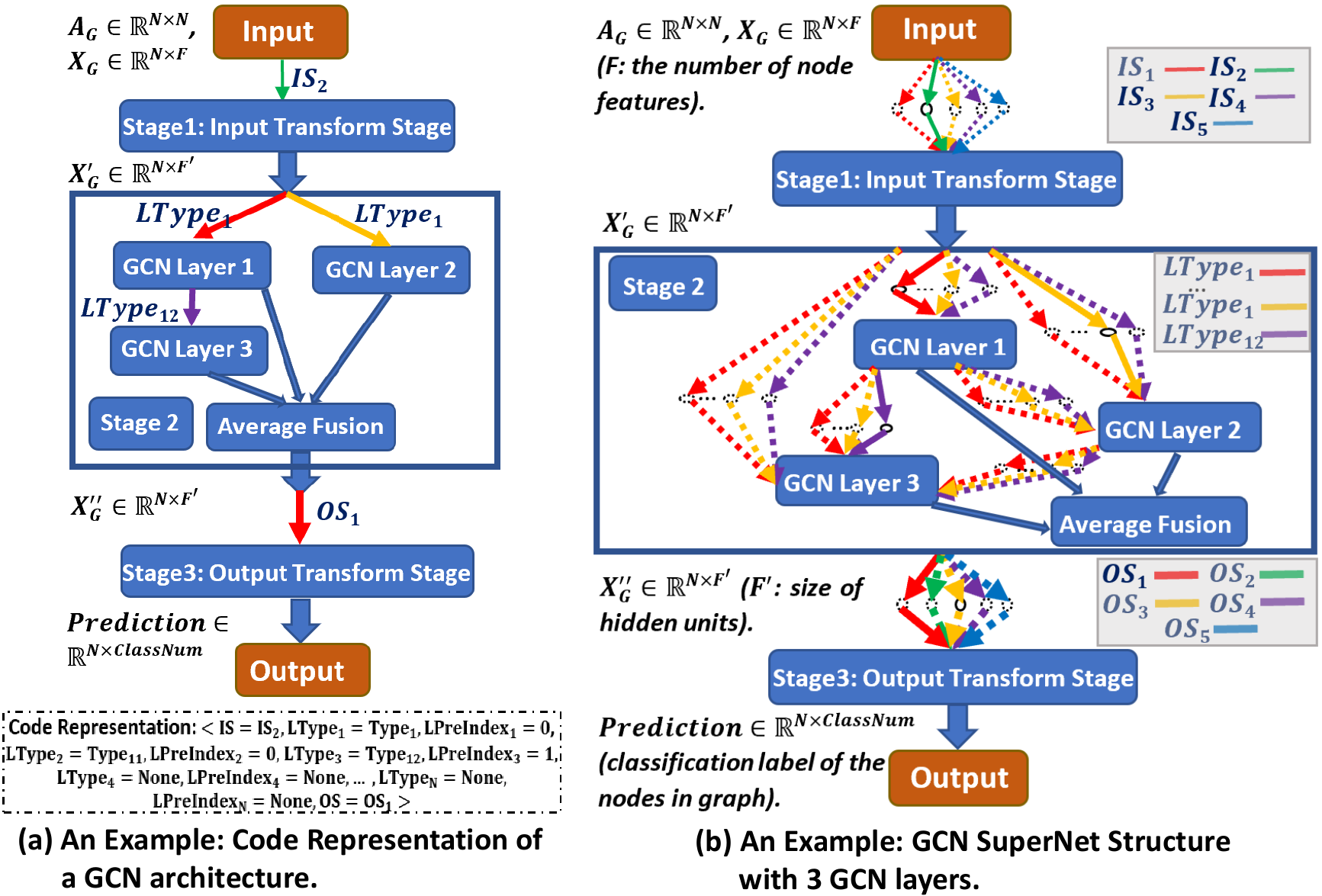}
	 \vspace{-0.6cm}
    \caption{Code Representation and GCN SuperNet Structure.}
\label{fig2}
\vspace{-0.4cm}
\end{figure}

Specifically, in Stage 1 and Stage 3, we use 2 parameters: Input Structure (IS) and Output Structure (OS), whose options are composed of fully connected layers with different activation functions, to specify the applied processing structure. For Stage 2, we utilize $LType_{i}$ and $LPreIndex_{i}$ to describe the operation type and connection method of each GCN layer $i (i=1,…,L)$, and apply the mean operation to fuse the outputs of all valid GCN layers. We take some state-of-the-art GCN layers implemented by the \textit{torch\_geometric} library\footnote{https://pytorch-geometric.readthedocs.io/en/latest/} as the options of $LType_{i}$, and allow each layer to connect with any one of its front layers, so as to build a flexible and diversified feature extraction architecture. Figure~\ref{fig2} (a) is an example of a GCN architecture code.

Considering the value space of each parameter, the above defined code representation can represent about $5\times \prod_{i=1}^{L} (12\times (i+1))\times 5=25\times 12^{L}\times (L+1)!$ GCN architectures in total. These architectures are regarded as the search spaces in FL-AGCNS, denoted as $\mathcal{A}$, and the code representation of a GCN architecture $a\in \mathcal{A}$ is denoted as $c_a$. 

\subsection{GCN SuperNet in FL-AGCNS}\label{section:3.2}

In FL-AGCNS, the main role of GCN SuperNet is to evaluate the performance of GCN architectures by sharing parameters for different architectures, thus efficiently complete the fitness evaluation in the evolution process. Specifically, given $a\in \mathcal{A}$, FL-AGCNS extracts corresponding weights from $\mathcal{W}$, i.e., the collection of all parameters in SuperNet, and thus efficiently transform the data flow to evaluate its performance. The realization of this function requires SuperNet to cover all possible GCN architectures in $\mathcal{A}$. Based on this thought, we design the structure of GCN SuperNet. Figure 2(b) is an example.

\textbf{Federated SuperNet Evaluation.} In FL-AGCNS, subgraphs are scattered in different clients without sharing. To obtain the performance of $a\in \mathcal{A}$ on such federated graph dataset, the local evaluation information of each client should be calculated separately, and therefore GCN SuperNet should be stored in each client. Let $\mathcal{W}_{a}=\mathcal{W}\odot c_{a}$ be the corresponding parameters of $a\in \mathcal{A}$, where $\odot$ is the mask operation that keeps parameters of the complete SuperNet only for positions corresponding to the operations applied in code $c_{a}$. Then, the performance of $a\in \mathcal{A}$ under the FL framework (we then call this the featured performance) can be expressed as follows:
\vspace{-0.1cm}
\begin{equation}
\small
FLL_{a}=\textbf{DE}\bigg[\sum_{i=1}^{N} \frac{|val_{G_i}|}{\sum_{j=1}^{N} |val_{G_j}|} \times \textbf{EN}\Big[\mathcal{L}_{val_{G_i}} (\mathcal{W}_{a},a)\Big]\bigg]
\vspace{-0.3cm}
\label{equ:1}
\end{equation}

where $\mathcal{L}_{val_{G_i}} (\mathcal{W}_{a},a)$ denotes the validation loss of architecture $a$ on subgraph $G_{i}$ under weight $\mathcal{W}_{a}$, \textbf{EN} and \textbf{DE} are homomorphic encryption and decryption method~\cite{p26} respectively. In FL-AGCN, each client calculates $\mathcal{L}_{val_{G_i}} (\mathcal{W}_{a},a)$ separately, then sends encrypted evaluation information to the controller to ensure the security of information transference. Controller aggregates local information and obtains the featured performance $FL\mathcal{L}_{a}$ after decryption. Detailed process is shown in the first round of information exchange in the right dotted box of Figure~\ref{fig1}.

\textbf{Federated SuperNet Optimization.} In addition to the evaluation task, SuperNet also needs to optimize $\mathcal{W}$ according to the federated performance of GCN architectures in population $\mathcal{P}$ provided by the optimizer, so as to achieve higher evaluation capability. Given $a\in \mathcal{P}$, the gradient of its architecture parameters $\mathcal{W}_{a}$ can be calculated as follows:
\vspace{-0.1cm}
\begin{equation}
\vspace{-0.7cm}
\small
\begin{split}
d\mathcal{W}_{a}=&\frac{\partial FL\mathcal{L}_{a}}{\partial \mathcal{W}_{a}}=\sum_{i=1}^{N} \frac{|train_{G_i}|}{\sum_{j=1}^{N} |train_{G_j}|} \times \frac{\partial\mathcal{L}_{train_{G_i}} (\mathcal{W}_{a},a)}{\partial\mathcal{W}_{a}}\\
&=\sum_{i=1}^{N} \frac{|train_{G_i}|}{\sum_{j=1}^{N} |train_{G_j}|}\times \frac{\partial\mathcal{L}_{train_{G_i}} (\mathcal{W}_{a},{a})}{\partial\mathcal{W}}\odot c_{a}
\end{split}
\vspace{-0.7cm}
\notag
\end{equation}

SuperNet parameters $\mathcal{W}$ should fit all the individuals in population $\mathcal{P}$, and thus the gradients for all architectures should be accumulated to calculate the gradient of $\mathcal{W}$:
\vspace{-0.1cm}
\begin{equation}
\vspace{-0.7cm}
\small
\begin{split}
d\mathcal{W}&=\frac{1}{|\mathcal{P}|} \sum_{a\in \mathcal{P}} d\mathcal{W}_{a} \\
&=\frac{1}{|\mathcal{P}|} \sum_{a\in \mathcal{P}} \sum_{i=1}^{N} \frac{|train_{Gi}|}{\sum_{j=1}^{N} |train_{G_j}|} \times \frac{\partial \mathcal{L}_{train_{Gi}} (\mathcal{W}_{a},a)}{\partial \mathcal{W}}\odot c_{a} \\
&=\frac{1}{|\mathcal{P}|} \sum_{i=1}^{N} \frac{|train_{Gi}|}{\sum_{j=1}^{N} |train_{G_j}|} \times \sum_{a\in \mathcal{P}} \frac{\partial \mathcal{L}_{train_{Gi}} (\mathcal{W}_{a},a)}{\partial \mathcal{W}}\odot c_{a}
\end{split}
\vspace{-0.7cm}
\notag
\end{equation}

Denote $\sum_{a\in \mathcal{P}} \frac{\partial \mathcal{L}_{train_{Gi}} (\mathcal{W}_{a},a)}{\partial \mathcal{W}}\odot c_{a}$, i.e., $d\mathcal{W}$ information of $\mathcal{P}$ obtained by client $i$, as $d\mathcal{W}_{G_i}^{\mathcal{P}}$, and add  homomorphic encryption to ensure the security of information transference, then the above equation can be written as:
\vspace{-0.1cm}
\begin{equation}
\vspace{-0.3cm}
\small
\begin{split}
d\mathcal{W}=\textbf{DE}\bigg[\frac{1}{|\mathcal{P}|} \sum_{i=1}^{N} \frac{|train_{G_i}|}{\sum_{j=1}^{N} |train_{G_j}|}\times \textbf{EN}\Big[d\mathcal{W}_{G_i}^{\mathcal{P}}\Big]\bigg]
\end{split}
\label{equ:2}
\vspace{-0.3cm}
\end{equation}

In the federated SuperNet optimization stage of FL-AGCNS, each client calculates $d\mathcal{W}_{G_i}^{\mathcal{P}}$ separately, then sends the encrypted gradient information to the controller. The controller aggregates the local information to obtain the encrypted $d\mathcal{W}$, i.e., $\textbf{EN}[d\mathcal{W}]$, and sends it to all clients. After decryption, each client can use $d\mathcal{W}$ to optimize the weight of $\mathcal{W}$ according to the gradient descent method, and thus improve the evaluation ability of GCN SuperNet. The operation flow of this part is shown in the left dotted box of Figure~\ref{fig1}.

\subsection{Federated Evolutionary Optimization}\label{section:3.3}

Architecture optimization is another core content of the HFL-based GCN NAS. In FL-AGCN, we introduce \underline{E}volutionary \underline{A}lgorithm (EA)~\cite{DBLP:books/mit/H1992} to optimize GCN architectures. We aim to recommend a GCN architecture that performs well on multiple subgraphs, so our optimization goal is to minimize the federated performance score mentioned in Section~\ref{section:3.2}. A simple EA implementation scheme is to take the negative of federated performance as the individual fitness, and execute evolution steps in the controller to optimize the population. 

\textbf{Two Difficulties.} However, this EA scheme, denoted as $EA_{simple}$, is not effective enough for GCN NAS tasks under the FL framework, having the following two defects. Firstly, it ignores the preference of each client during evolution, and thus be less efficient. Specifically, we find that individuals in the population may not be applicable to some clients in the early evolution, and thus show poor federated performance in FL framework. If GCN architectures preferred by each client can be involved in the population evolution, they can guide individuals to efficiently cover their shortages, and thus speed up the optimization process. However, $EA_{simple}$ scheme fails to do so and thus be inefficient.

Secondly, due to lacks of dominant architectures of clients, the population in $EA_{simple}$ may lead to the unfairness of SuperNet evaluation, and thus falls into the local optimal solution. Specifically, in each evolution step, our algorithm firstly uses population to optimize SuperNet parameters, and then utilizes the optimized SuperNet to evaluate individual fitness and evolves a new population. In such a scenario, the population's performance skew would greatly affect the fairness of SuperNet evaluation. That is to say, if the federated performance of population is heavily skewed towards partial clients (denoted by $Clients_{Apply}$), SuperNet weights will show performance evaluation deviation after optimization, i.e., score high on architectures suitable for $Clients_{Apply}$ while ignoring other subgraphs. With a long time, $EA_{simple}$ would fall into local optimums under the wrong guidance of SuperNet. A solution to the performance skew problem is to add dominant architectures of each client to population to optimize SuperNet parameters, and thus enhance the fairness of evaluation. However, the $EA_{simple}$ fails to do so. The federated performance of its population is generally low at the early stage, which may lead to performance skew, resulting in poor optimization effect.

\begin{algorithm}[t]
	\small
	\caption{Federated Evolutionary Optimization}
		\label{algo:algo1}
		\centering
        \flushleft{\textit{\textbf{\underline{Client i (i=1,...,N)}}}}
	    \begin{algorithmic}[1]
		    \STATE $\textbf{EN}[\mathcal{L}_{val_{G_i}}(\mathcal{P})]\gets\{\textbf{EN}[\mathcal{L}_{val_{G_i}}(W_a,a)] |a \in \mathcal{P}\}$, send it to Controller.
		    \STATE Receive $\mathcal{P}_{FL}$ from Controller
		    \STATE $\mathcal{P}_{G_i} \gets \mathcal{P}_{G_i} \cup \mathcal{P}_{FL}$, $\mathcal{L}_{val_{G_i}}(\mathcal{P}) \gets \{\mathcal{L}_{val_{G_i}}(W_a,a)|a \in \mathcal{P}_{G_i}\}$
		    \STATE $\mathcal{P}_{G_i} \gets$ Consider $-\mathcal{L}_{val_{G_i}}(\mathcal{P})$ as fitness values of $\mathcal{P}_{G_i}$, and execute evolutionary operations to optimize $\mathcal{P}_{G_i}$.
		    \STATE $\mathcal{P}_{G_i}' \gets$ TOP $(|\mathcal{P}|\times \gamma\times\sum_{i=1}^{N}{\frac{|val_{G_i}|}{\sum_{j = 1}^{N}{|val_{G_j}|}}})$ individuals with lowest $\mathcal{L}_{val_{G_i}}$ values in $\mathcal{P}_{G_i}$.
		    \STATE 	Send $\mathcal{P}_{G_i}'$ and $\textbf{EN}[\mathcal{L}_{val_{G_i}}(\mathcal{P}_{FL})]$ to Controller.
	    \end{algorithmic}
       \flushleft{\textit{\textbf{\underline{Controller}}}}
	    \begin{algorithmic}[1]
		    \STATE Receive the encrypted $\mathcal{L}_{val_{G_i}}$ of $\mathcal{P}$ from Clients.
		    \STATE $FL\mathcal{L}(\mathcal{P})\gets \{FL\mathcal{L}_a|a\in \mathcal{P}\}$
		    \STATE $\mathcal{P}_{FL} \gets$ Consider $-$$FL\mathcal{L}(\mathcal{P})$ as fitness values of $\mathcal{P}$, execute evolutionary operations to optimize $\mathcal{P}$. Send $\mathcal{P}_{FL}$ to Clients.
		    \STATE 	Receive $\mathcal{P}_{G_i}'$ and the encrypted $\mathcal{L}_{val_{G_i}}(\mathcal{P}_{FL})$ from Clients.
		    \STATE 	$\mathcal{P}_{FL}' \gets$ TOP $(|\mathcal{P}|\times(1-\gamma))$ individuals with lowest $FL\mathcal{L}$ values in $\mathcal{P}_{FL}$.
		    \STATE $\mathcal{P} \gets \bigcup_{i=1}^{N}{\mathcal{P}_{G_i}'}\cup \mathcal{P}_{FL}'$
	    \end{algorithmic}
\end{algorithm}

\textbf{FEO strategy.} To better solve the HFL based GCN NAS problem, in this paper, we propose a \underline{F}ederated \underline{E}volutionary \underline{O}ptimization strategy (FEO strategy), which fully considers the preference of each client during the evolution for optimization acceleration and optimization effect improvement. In FEO strategy, each client and the controller maintain an EA optimizer, so as to search for the population with good federated performance, which is denoted as $\mathcal{P}_{FL}$, and the population performing well in a specific subgraph $G_i$, which is denoted as $\mathcal{P}_{G_i}$, respectively\footnote{Controller's optimizer takes the negative of federated performance $FL\mathcal{L}_{a}$ as the individual fitness, aiming to improve the federated performance of the population. The EA optimizer in each client $i$ uses $\mathcal{L}_{val_{G_i}} (\mathcal{W}_a,a)$ to measure individual fitness instead, aims to explore characteristics of subgraph $G_i$ and get GCN architectures suitable for $G_i$.}. In each round of evolution, FEO strategy firstly evolves new $\mathcal{P}_{FL}$ and $\mathcal{P}_{G_i}$  (i=1,…,N), and then optimizes SuperNet parameters using a new population $\mathcal{P}$, which is composed of top $|\mathcal{P}|\times \gamma \times \sum_{i=1}^{N} \frac{|val_{G_i}|}{\sum_{j=1}^{N} |val_{G_j}|}$ individuals with lowest $\mathcal{L}_{val_{G_i}}$ scores in each $\mathcal{P}_{G_i}$ and top $|\mathcal{P}|\times (1-\gamma)$  individuals with the lowest $FL\mathcal{L}$ scores in $\mathcal{P}$, where $0<\gamma<1$ is the proportion of clients' evolutionary results in $\mathcal{P}$. Considering the GCN architectures favored by each client, new population $\mathcal{P}$ can effectively avoid the performance skew problem, maintaining the fairness of GCN SuperNet evaluation.

After the SuperNet optimization step, FEO strategy replaces $\mathcal{P}_{FL}$ with the new population $\mathcal{P}$ to perform the next round of evolution. For that individuals in $\mathcal{P}_{G_i}$ are involved in the population evolution of controller, the shortages of individuals can be efficiently corrected in the next round of evolution, achieving higher federated performance. After adding the local EA operations, FEO strategy successfully overcomes two difficulties of $EA_{simple}$, performing well on GCN NAS problems under FL framework. The right dotted box of Figure~\ref{fig1} shows the detailed procedure of FEO, and Algorithm~\ref{algo:algo1} gives its pseudo-code.

Note that in FEO strategy, the value of $\gamma$ will decrease with the increase of evolution generations. The reasons are as follows. With the increase of generations, the shortages of individuals in $\mathcal{P}$ are gradually resolved and their federated performance are improved, the performance skew phenomenon disappears progressively, and SuperNet parameters are gradually stabilized. In this case, the advantages of introducing the preference of each client will gradually decrease. Promoting the proportion of $\mathcal{P}_{FL}$ in $\mathcal{P}$ can make the optimizer focus on more individuals with high federated performance, and turn to apply the advantage fusion strategy to evolve individuals with better federated performance.

\begin{algorithm}[t]
	\small
	\caption{FL-AGCN Algorithm}
		\label{algo:FL-AGCN_Algorithm}
		\centering
	    \begin{algorithmic}[1]
		    \STATE \textbf{\# Initialization}
		    \STATE  \textit{Controller:} Initialize the weights of GCN SuperNet, population $\mathcal{P}$. Send them to N Clients.
		    \STATE \textit{Client i (i=1,...,N):} Receive $\mathcal{P}$ and SuperNet weights from Controller. Initialize SuperNet and local population $\mathcal{P}_{G_i}$$\gets$$\mathcal{P}$.
		    \STATE \textbf{\# Searching for the Optimal GCN architecture}
		    \FOR {$t = 1, ... , E_{evo}$}
		        \STATE \textbf{\# Federated SuperNet Weights Optimization}
		        \STATE \textit{Controller:} send $\mathcal{P}$ to N Clients.
		        \STATE \textit{Client i (i=1,...,N):} Receive $\mathcal{P}$ from Controller.
		        \FOR{$w=1,...,E_{weight}$}
		            \STATE \textit{Client i (i=1,...,N):} Send $\textbf{EN}[d\mathcal{W}_{G_i}^{\mathcal{P}}]$ to Controller.
		            \STATE \textit{Controller:} Receive gradient information from N Clients. Get $d\mathcal{W}$ using Equation~\ref{equ:2}, send it to N Clients.
		            \STATE \textit{Client i (i=1,...,N):} Receive $d\mathcal{W}$ from Controller. Update the weights of SuperNet using gradient descent.
		        \ENDFOR
	            \STATE \textbf{\# Execute FEO Strategy}
	            \STATE \textit{Controller:} $\gamma=0.5\times0.99^t$, send $\gamma$ to N Clients.
	            \STATE \textit{Controller\&Clients:} Execute Algorithm~\ref{algo:algo1}.
		    \ENDFOR
		    \STATE \textbf{\# Output the Optimal GCN architecture}
		    \STATE \textit{Controller:} send $\mathcal{P}$ to N clients;
	        \STATE \textit{Client i (i=1,...,N):} Receive $\mathcal{P}$ from Controller. Calculate $\textbf{EN}[\mathcal{L}_{val_{G_i}}(\mathcal{W}_a,a)](a\in \mathcal{P})$ and send it to Controller.
	        \STATE \textit{Controller:} Receive $\textbf{EN}[\mathcal{L}_{val_{G_i}}(\mathcal{W}_a,a)] (a\in \mathcal{P})$ from N Clients. Get $FL\mathcal{L}_a (a\in \mathcal{P})$ according to Equation~\ref{equ:1}.
	        \STATE \textit{Controller:} $OPT_{GCN} \gets \mathop{argmin}_{a\in \mathcal{P}} FL\mathcal{L}_{a}$
		   \STATE {\bfseries Output: $OPT_{GCN}$}
        \end{algorithmic}
\end{algorithm}

\subsection{FL-AGCN: Federated Learning Framework for Automatic Graph Convolutional Network Search}\label{section:3.4}


Combining the GCN SuperNet with the FEO strategy, then we get the FL-AGCN algorithm. Algorithm~\ref{algo:FL-AGCN_Algorithm} summarizes the detailed procedure of FL-AGCN. Given a HFL based GCN NAS task, FL-AGCN firstly performs initialization steps (Line 1-3). Then it performs evolutionary steps iteratively to search for optimal GCN architecture (Line 4-15). In each evolution step, FL-AGCN synchronously updates SuperNet parameters in each client (Line 6-12), and executes FEO strategy to optimize population (Line 13-15). Finally, FL-AGCN selects an individual with the highest federated performance as the final output (Line 16-20). Search time analysis of FL-AGCN is given in supplementary material.

\begin{table}[t]
    \centering
    \caption{Datasets used in our experiment.}
\resizebox{0.5\textwidth}{!}{
    \begin{tabular}{lccccc}
        \toprule
        Dataset & \#Nodes & \#Edges & \#Classes & \#Features & Train/Dev/Test Nodes\\
        \midrule
        Cora & 2,708 & 5,429 & 7 & 1433 & 140/500/1,000\\
        CiteSeer & 3,327 & 4,732 & 6 & 3,703 & 120/500/1,000\\
        PubMed & 19,717 & 44,338 & 3 & 500 & 60/500/1,000\\
        CoraFull & 19,793 & 126,842 & 70 & 8,710 & 1,395/500/1,000\\
        Physics & 34,493 & 495,924 & 5 & 8,415 & 100/500/1,000\\
        \bottomrule
    \end{tabular}
}
\vspace{-0.3cm}
    \label{tab:dataset}
\end{table}

\begin{table*}[h]
\centering
\caption{Experimental results on three datasets: Cora, CiteSeer and PubMed.}
\begin{small}
\resizebox{0.99\textwidth}{!}{
\begin{tabular}{lccccccccccc}
\toprule
\multirow{2}{*}{} & \multicolumn{3}{c}{\textbf{Cora}} && \multicolumn{3}{c}{\textbf{CiteSeer}} && \multicolumn{3}{c}{\textbf{PubMed}}\\
\cline{2-4}\cline{6-8}\cline{10-12}
&$\mathcal{FLACC}$(\%)&Params(M)&Time(s)&&$\mathcal{FLACC}$(\%)&Params(M)&Time(s)&&$\mathcal{FLACC}(\%)$&Params(M)&Time(s)\\
\midrule
FL-AGCNS&\textbf{82.9}&0.105 &0.0339 &&\textbf{70.6} &0.250 &0.0553 &&\textbf{78.4} & 0.036&0.0601 \\
FL-Gradient&65.4 &0.045 &0.0415 &&37.0 &0.300 &0.6061 &&77.9 & 0.099&0.0720 \\
FL-RL&75.6 &0.101 &0.0297 &&\underline{70.2} &0.242 &0.0631 &&68.6 &0.032 &0.0373 \\
FL-Random&79.8 &0.113 &0.0335 &&47.4 &0.296 &0.0916 &&64.3 &0.066 &0.0492 \\
\midrule
GAT&79.0 & 0.092& 0.0262&& 68.0& 0.238& 0.0612&& 77.5& 0.032& 0.0502\\
SAGE&78.8 &0.184 & 0.0294&&66.6 & 0.475& 0.0920&&75.2 &0.064 & 0.0426\\
GCN&81.3 &0.102 &0.0200 &&68.6 &0.260 &0.0467 &&77.3 &0.034 &0.0395 \\
SGC&78.1 &0.010 &0.0152 &&67.0 &0.022 &0.0381 &&76.3 & 0.002 &0.0417 \\
APPNP&\underline{82.2} &0.092 & 0.0216&&69.5 &0.237 & 0.0572&&77.5 &0.032 &0.0401 \\
AGNN&81.2 &0.023 &0.0820 &&68.7 &0.059 &0.0969 &&\underline{78.1} &0.008 & 0.0960\\
ARMA&75.2 &0.092 &0.0246 &&63.8 &0.237 &0.0462 &&76.8 &0.032 &0.0360 \\
GatedGraph&76.6 &0.130 &0.0270 &&63.5 &0.275 &0.0494 &&77.3 &0.070 &0.0361 \\
\bottomrule
\end{tabular}
}
\vspace{-0.4cm}
\end{small}
\label{tab:result_on_small}
\end{table*}

\begin{table}[h]
\centering
\caption{Results on the other two datasets: CoraFull and Physics.}
\resizebox{0.5\textwidth}{!}{
\begin{tabular}{lccccccc}
\toprule
\multirow{2}{*}{\textbf{}} & \multicolumn{3}{c}{\textbf{CoraFull}} && \multicolumn{3}{c}{\textbf{Physics}}\\
\cline{2-4}\cline{6-8}
&$\mathcal{FLACC}$(\%)&Params(M)&Time(s)&&$\mathcal{FLACC}$(\%)&Params(M)&Time(s)\\
\midrule
FL-AGCNS&\underline{59.7} &0.562 &0.4859 &&\textbf{77.1} &0.547 &0.4012 \\
FL-Gradient& 58.0&0.450 &0.6381 &&67.0 &0.539 &0.4162 \\
FL-RL&44.2 &0.562 &0.4813 &&61.8 &0.539 &0.3722 \\
FL-Random&50.0 &0.291 &0.5815 &&72.8 &0.576 &0.3778 \\
\midrule
GAT     &57.5   &0.562    &0.6381     &&64.9&0.539&0.5399\\
SAGE    &57.3   &1.124  &0.6385     &&70.1&1.078&0.9268\\
GCN     &51.5   &1.172  &0.7674     &&\underline{76.5}&0.581&0.5130\\
SGC     &56.1   &0.610    &0.5420     &&70.2&0.042&0.3460\\
APPNP   &\textbf{60.3}   &0.562    &0.4773     &&73.2&0.539&0.6709\\
AGNN    &47.2   &0.141    &0.5188     &&69.9&0.135&0.3399\\
ARMA    &47.0   &0.562    &0.5506     &&58.8&0.539&0.5171\\
GatedGraph&31.7 &0.599    &0.4726     &&62.8&0.576&0.4052\\
\bottomrule
\end{tabular}
}
\vspace{-0.4cm}
\label{tab:result_on_large}
\end{table}

\section{Experiment}\label{section:4}

In this section, we test the performance of FL-AGCNS. Firstly, we compare FL-AGCNS with existing NAS algorithms. Secondly, we compare the optimal GCN architecture discovered by FL-AGCNS with the state-of-the-art GCN architectures. Finally, ablation experiment is conducted to analyze FEO strategy designed in our algorithm. All experiments are implemented using RTX 2080 Ti.

\subsection{Experimental settings}\label{section:4.1}

\textbf{Datasets.} We evaluate the proposed algorithm on 5 popular network datasets: Cora, CiteSeer, PubMed~\cite{DBLP:journals/corr/YangCS16}, CoraFull~\cite{DBLP:journals/corr/BojchevskiG17} and Physics~\cite{DBLP:journals/corr/abs-1811-05868}. The first four datasets are citation networks, and the last dataset is a co-author network. The dataset statistics are given in Table 2. 

\textbf{Evaluation Metrics.} We use the federated accuracy $\mathcal{FLACC}(a)$ which is defined as follows to examine the performance of GCN architecture $\alpha\in \mathcal{A}$ under FL framework.
\vspace{-0.3cm}
\begin{equation}
\vspace{-0.3cm}
\small
\mathcal{FLACC}(\alpha) = \sum_{i=1}^{N} \frac{|test_{G_i}|}{\sum\limits_{j=1}^{N} |test_{G_j}|} \times \mathcal{ACC}_{test_{G_i}}(w^{\ast}(\alpha),\alpha)
\notag
\vspace{-0.3cm}
\end{equation}
\begin{equation}
\small
s.t.\ w^{\ast}(\alpha)=\mathop{argmin}\limits_{w} \sum_{i=1}^{N} \frac{|train_{G_i}|}{\sum\limits_{j=1}^{N} |train_{G_j}|}\times \mathcal{L}_{train_{G_i}}(w,\alpha)
\notag
\vspace{-0.2cm}
\end{equation}
where $\mathcal{ACC}_{test_{G_i}} (w^{\ast}(\alpha),\alpha)$ denotes the test accuracy of $\alpha$ on subgraph $G_{i}$ under weight $w^{\ast}(\alpha)$. For each GCN NAS algorithm, we report the $\mathcal{FLACC}$ score of the optimal GCN architecture discovered by it, so as to examine its ability in dealing with the HFL based GCN NAS problems. We also report GCN architectures' parameter amount (denoted by Params) and the inference time on the test set (denoted by Time), to show their complexity.


\textbf{Baselines.} We compare FL-AGCNS with two popular NAS algorithms: reinforcement learning based BlockQNN~\cite{BlockQNN} and gradient based DARTS~\cite{p23}, and a commonly used baseline in NAS, Random Search. To enable these non-federated NAS algorithms to cope with our HFL based GCN NAS problems, we set the search space to $\mathcal{A}$, and replace their evaluation or gradient information with the federated ones. We then denote these federated versions as FL-RL, FL-Gradient and FL-Random respectively. In FL-RL and FL-Random, we evalue each GCN architecture for 50 epochs. In addition, we take 8 state-of-the-art GNN architectures: GAT~\cite{p13}, SAGE~\cite{p15}, GCN~\cite{p9}, SGC~\cite{p11}, APPNP~\cite{p10}, AGNN~\cite{agnn}, ARMA~\cite{arma} and GatedGraph~\cite{GatedGraph}, as baselines, to show the importance of GCN NAS under FL framework.


\textbf{Implementation Details.} In the experiments, we divide each network into $N=3$ subgraphs using Metis~\cite{Metis} clustering method, and use these subgraphs to simulate $N=3$ clients in FL framework\footnote{In the experiment, we simulate $N=8$ clients instead in the Physics dataset, due to GPU memory constraints.}. We apply CKKS scheme\footnote{https://github.com/muhanzhang/SEAL/tree/master/Python}~\cite{p26} to achieve encryption and decryption. In FL-AGCNS, we set population size to 60, the number of GCN layers $L$ to 6, evolution generations $E_{evo}$ to 250 and $E_{weight}$ to 5. As for the compared NAS algorithms and existing GCN architectures, we follow implementation details reported in their papers, and control the running time of each NAS algorithm to be the same.



\begin{figure}[t]
    \centering
    \includegraphics[width = 0.99\linewidth]{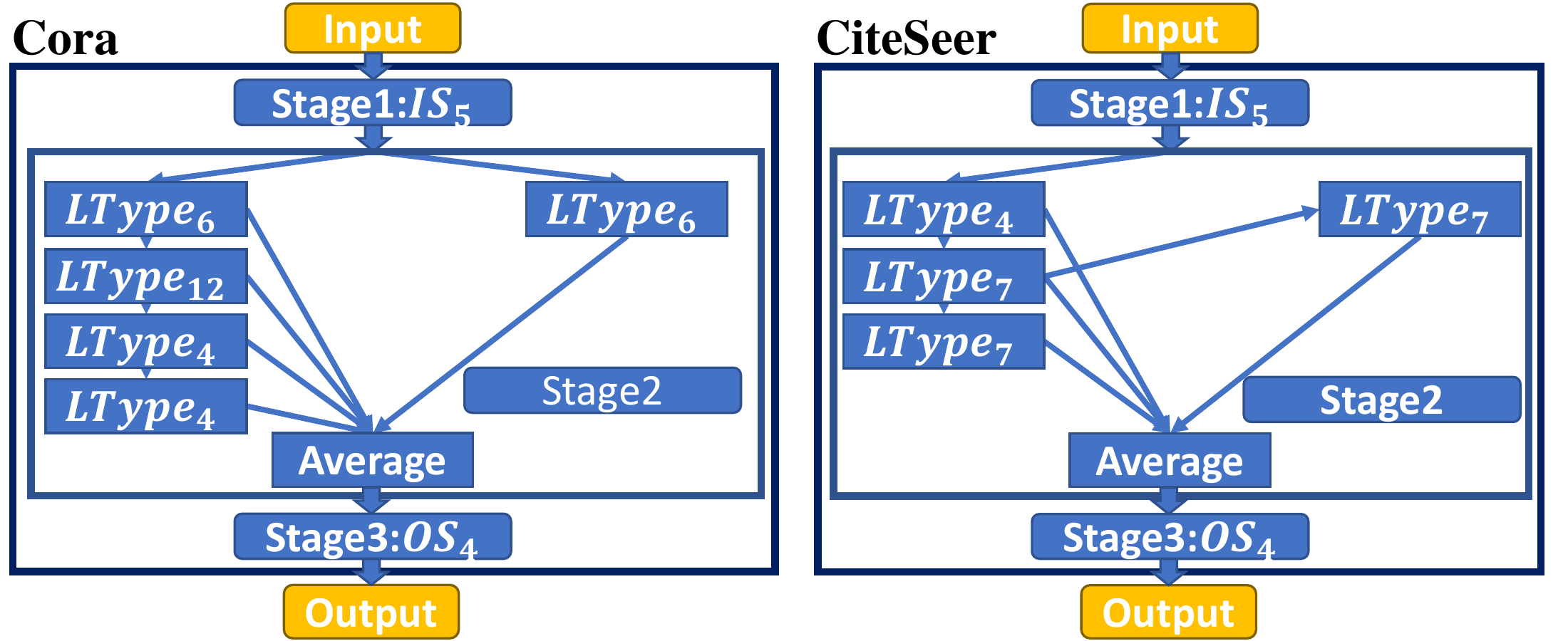}
	 \vspace{-0.3cm}
	 \caption{Optimal GCN architecture searched by FL-AGCNS.}
    \label{fig:network}
\vspace{-0.5cm}
\end{figure}

\begin{figure*}[h]
    \centering
    \includegraphics[width=0.99\linewidth]{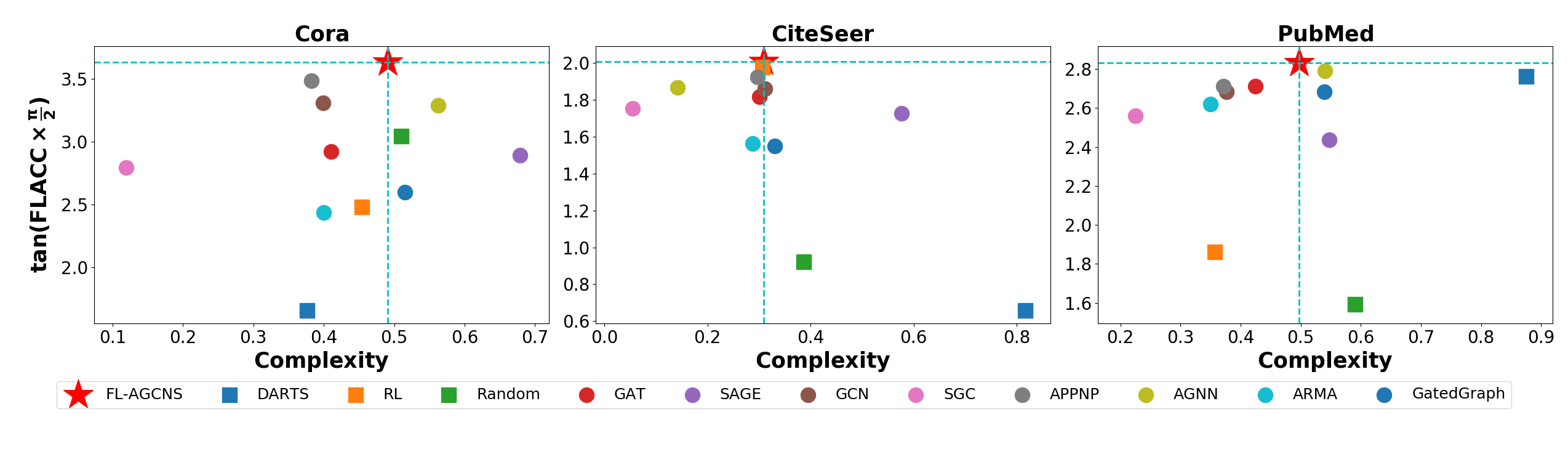}
	 \vspace{-0.6cm}
    \caption{Visualization of experiment results. $Complexity(a)=\frac{1}{2}\times (\frac{Params_{a}}{\max_{a'\in AList}Params_{a'}}+\frac{Time_{a}}{\max_{a'\in AList}Time_{a'}})$ represents the complexity of the GCN architecture $a\in AList$, where $AList$ denotes 12 architectures determined by algorithms or models in Table 3.}
    \label{fig:analyse}
\vspace{-0.3cm}
\end{figure*}

\subsection{Performance Evaluation}\label{section:4.2}

The performance of 4 NAS algorithms and 8 GCN architectures under FL framework are shown in Table 3 and Table 4, Figure 3 visualizes GCN architectures searched by FL-AGCNS. We can observe that FL-AGCNS exceeds the other NAS algorithms on five datasets, achieving higher performance in the FL framework. This result shows the importance of local optimizers under FL scenarios. With the help of local optimizers, FL-AGCNS can utilize the preference of each client to quickly enhance population's federated performance under FL scenarios, thus recommends more powerful GCN architectures within the same search time, compared with other NAS methods without local optimizers.

In addition, we observe that GCN architectures discovered by FL-AGCNS generally outperform existing GCN architectures in FL framework. This result shows us the importance of GCN NAS. GCN NAS algorithms make FL more flexible and powerful. They can provide clients with more and better GCN models in the FL framework, achieving stronger federated effect, compared with the existing models.

Figure 4 visualizes GCN architectures' performance and complexity in Table 3. As we can see, GCN architectures discovered by FL-AGCNS achieve the highest performance scores, and their complexity exceeds many of the existing classic GCN architectures and NAS solutions. Overall, the solution provided by FL-AGCNS is superior considering both performance and complexity.




\subsection{Ablation Experiments}\label{section:4.3}

We further investigate the effect of FEO strategy and connections among each clients on the performance of FL-AGCNS algorithm using the following three variants of FL-AGCNS, thus verify the innovations presented in this paper and specify the application conditions of our FL-AGCNS.
\begin{itemize}
\vspace{-0.2cm}
\item [1.] CrotrollerEO: This algorithm fixes the value of $\gamma$ in FL-AGCNS as 1. It only preserves the EA optimizer in the controller.
\vspace{-0.1cm}
\item [2.] ClientEO: This algorithm fixes the value of $\gamma$ in FL-AGCNS as 0. It only preserves the local EA optimizer in each client.
\vspace{-0.1cm}
\item [3.] RandomPartition: This algorithm uses randomly-divided subgraphs to simulate clients in the FL framework, and executes FL-AGCNS on those clients.
\vspace{-0.2cm}
\end{itemize}


Corresponding results are shown in Table 5, we can see that FL-AGCNS has much better performance than CrotrollerEO and ClientEO. This result shows us the significance and necessity of maintaining EA optimizers in both the controller and clients under FL scenarios. As discussed in Section~\ref{section:3.3}, EA optimizers in each client contribute much in avoiding local optimums and improving federated performance in the early generations, whereas the controller's EA optimizer performs better on population optimization in the later generations. Two kinds of optimizers optimize population from different aspects in FL-AGCNS and show good performance in different stages. Therefore, their combination is meaningful and necessary.

Besides, we observe that RandomPartition performs much worse than FL-AGCNS. We notice that the random partition method generates much more external connections among subgraphs compared with the Metis method, which leads to many useful edge information being ignored in the FL framework. This result shows that FL-AGCNS is more suitable for FL scenarios where the subgraph of each client is densely connected and its outer connections are of no great importance and rare in practise. For example, the clients contain graph information of different districts or different user groups.

\begin{table}[t]
\centering
\caption{Results of ablation experiments on Cora dataset.}
\resizebox{0.96\columnwidth}{!}{
\begin{tabular}{lccc}
\toprule
Algorithm&$\mathcal{FLACC}$(\%)&Params(M)&Time(s)\\
\midrule
FL-AGCNS&82.9&0.105 &.0339  \\
ControllerEO& 71.6& 0.117&  .0434\\
ClientEO& 78.6& 0.105&.0489 \\
\midrule
RandomPartition&52.6 &0.143 &.0526\\
\bottomrule
\end{tabular}
}
\vspace{-0.3cm}
\label{tab:result_of_ablation}
\end{table}

\section{Conclusion and Future Works}\label{section:5}

In this paper, we make up for the shortcomings of the existing GCN NAS techniques, and propose FL-AGCNS, a more efficient and practical GCN NAS algorithm which considers the FL scenarios. Our algorithm designs the FEO strategy to fully consider the preferences of different clients, enabling distributed agents to collaboratively design the high-performance GCN model while protecting data privacy. As we know, this is the first work which combines GCN NAS with the FL. In addition, we apply the GCN SuperNet to reduce the evaluation cost of GCN models, and thus promote the search efficiency of FL-AGCNS. The extensive experiments show that our FL-AGCNS can recommend better GCN models in short time under the FL framework, surpassing the state-of-the-arts NAS methods and GCN models, which demonstrates the effectiveness of FL-AGCNS. In this paper, we only consider one optimization objective and the horizontal FL scenarios. In future works, we will consider more optimization objectives and more FL scenarios, so as to recommend more practical GCN models under more realistic scenarios.

\nocite{langley00}


\begin{thebibliography}{32}
\providecommand{\natexlab}[1]{#1}
\providecommand{\url}[1]{\texttt{#1}}
\expandafter\ifx\csname urlstyle\endcsname\relax
  \providecommand{\doi}[1]{doi: #1}\else
  \providecommand{\doi}{doi: \begingroup \urlstyle{rm}\Url}\fi

\bibitem[Bianchi et~al.(2019)Bianchi, Grattarola, Livi, and Alippi]{arma}
Bianchi, F.~M., Grattarola, D., Livi, L., and Alippi, C.
\newblock Graph neural networks with convolutional {ARMA} filters.
\newblock \emph{CoRR}, abs/1901.01343, 2019.
\newblock URL \url{http://arxiv.org/abs/1901.01343}.

\bibitem[Bojchevski \& G{\"{u}}nnemann(2017)Bojchevski and
  G{\"{u}}nnemann]{DBLP:journals/corr/BojchevskiG17}
Bojchevski, A. and G{\"{u}}nnemann, S.
\newblock Deep gaussian embedding of attributed graphs: Unsupervised inductive
  learning via ranking.
\newblock \emph{CoRR}, abs/1707.03815, 2017.

\bibitem[Cheon et~al.(2017)Cheon, Kim, Kim, and Song]{p26}
Cheon, J.~H., Kim, A., Kim, M., and Song, Y.~S.
\newblock Homomorphic encryption for arithmetic of approximate numbers.
\newblock In Takagi, T. and Peyrin, T. (eds.), \emph{ASIACRYPT}, volume 10624
  of \emph{Lecture Notes in Computer Science}, pp.\  409--437. Springer, 2017.

\bibitem[Dong \& Yang(2019)Dong and Yang]{p24}
Dong, X. and Yang, Y.
\newblock Searching for a robust neural architecture in four {GPU} hours.
\newblock In \emph{CVPR}, pp.\  1761--1770. Computer Vision Foundation /
  {IEEE}, 2019.

\bibitem[Gao et~al.(2020)Gao, Yang, Zhang, Zhou, and Hu]{p5}
Gao, Y., Yang, H., Zhang, P., Zhou, C., and Hu, Y.
\newblock Graph neural architecture search.
\newblock In Bessiere, C. (ed.), \emph{IJCAI}, pp.\  1403--1409. ijcai.org,
  2020.

\bibitem[Hamilton et~al.(2017)Hamilton, Ying, and Leskovec]{p15}
Hamilton, W.~L., Ying, Z., and Leskovec, J.
\newblock Inductive representation learning on large graphs.
\newblock In Guyon, I., von Luxburg, U., Bengio, S., Wallach, H.~M., Fergus,
  R., Vishwanathan, S. V.~N., and Garnett, R. (eds.), \emph{Annual Conference
  on Neural Information Processing Systems}, pp.\  1024--1034, 2017.

\bibitem[Holland(1992)]{DBLP:books/mit/H1992}
Holland, J.~H.
\newblock \emph{Adaptation in Natural and Artificial Systems: An Introductory
  Analysis with Applications to Biology, Control, and Artificial Intelligence}.
\newblock {MIT} Press, 1992.
\newblock ISBN 9780262275552.
\newblock \doi{10.7551/mitpress/1090.001.0001}.
\newblock URL \url{https://doi.org/10.7551/mitpress/1090.001.0001}.

\bibitem[Hu et~al.(2019)Hu, Zhu, Wu, Wang, and Tan]{p1}
Hu, F., Zhu, Y., Wu, S., Wang, L., and Tan, T.
\newblock Hierarchical graph convolutional networks for semi-supervised node
  classification.
\newblock In Kraus, S. (ed.), \emph{IJCAI}, pp.\  4532--4539. ijcai.org, 2019.

\bibitem[Karypis \& Kumar(1998)Karypis and Kumar]{Metis}
Karypis, G. and Kumar, V.
\newblock A fast and high quality multilevel scheme for partitioning irregular
  graphs.
\newblock \emph{{SIAM} J. Sci. Comput.}, 20\penalty0 (1):\penalty0 359--392,
  1998.

\bibitem[Kipf \& Welling(2017)Kipf and Welling]{p9}
Kipf, T.~N. and Welling, M.
\newblock Semi-supervised classification with graph convolutional networks.
\newblock In \emph{ICLR}. OpenReview.net, 2017.

\bibitem[Klicpera et~al.(2019)Klicpera, Bojchevski, and G{\"{u}}nnemann]{p10}
Klicpera, J., Bojchevski, A., and G{\"{u}}nnemann, S.
\newblock Predict then propagate: Graph neural networks meet personalized
  pagerank.
\newblock In \emph{ICLR}. OpenReview.net, 2019.

\bibitem[Li et~al.(2016)Li, Tarlow, Brockschmidt, and Zemel]{GatedGraph}
Li, Y., Tarlow, D., Brockschmidt, M., and Zemel, R.~S.
\newblock Gated graph sequence neural networks.
\newblock In Bengio, Y. and LeCun, Y. (eds.), \emph{ICLR}, 2016.

\bibitem[Liu et~al.(2019)Liu, Simonyan, and Yang]{p23}
Liu, H., Simonyan, K., and Yang, Y.
\newblock {DARTS:} differentiable architecture search.
\newblock In \emph{ICLR}. OpenReview.net, 2019.

\bibitem[Liu et~al.(2020)Liu, Huang, Luo, Huang, Liu, Chen, Feng, Chen, Yu, and
  Yang]{p19}
Liu, Y., Huang, A., Luo, Y., Huang, H., Liu, Y., Chen, Y., Feng, L., Chen, T.,
  Yu, H., and Yang, Q.
\newblock Fedvision: An online visual object detection platform powered by
  federated learning.
\newblock In \emph{AAAI}, pp.\  13172--13179. {AAAI} Press, 2020.

\bibitem[McMahan et~al.(2017)McMahan, Moore, Ramage, Hampson, and y~Arcas]{p17}
McMahan, B., Moore, E., Ramage, D., Hampson, S., and y~Arcas, B.~A.
\newblock Communication-efficient learning of deep networks from decentralized
  data.
\newblock In Singh, A. and Zhu, X.~J. (eds.), \emph{AISTATS}, volume~54 of
  \emph{Proceedings of Machine Learning Research}, pp.\  1273--1282. {PMLR},
  2017.

\bibitem[McMahan et~al.(2016)McMahan, Moore, Ramage, and y~Arcas]{p21}
McMahan, H.~B., Moore, E., Ramage, D., and y~Arcas, B.~A.
\newblock Federated learning of deep networks using model averaging.
\newblock \emph{CoRR}, abs/1602.05629, 2016.

\bibitem[Shchur et~al.(2018)Shchur, Mumme, Bojchevski, and
  G{\"{u}}nnemann]{DBLP:journals/corr/abs-1811-05868}
Shchur, O., Mumme, M., Bojchevski, A., and G{\"{u}}nnemann, S.
\newblock Pitfalls of graph neural network evaluation.
\newblock \emph{CoRR}, abs/1811.05868, 2018.
\newblock URL \url{http://arxiv.org/abs/1811.05868}.

\bibitem[Shuman et~al.(2013)Shuman, Narang, Frossard, Ortega, and
  Vandergheynst]{p12}
Shuman, D.~I., Narang, S.~K., Frossard, P., Ortega, A., and Vandergheynst, P.
\newblock The emerging field of signal processing on graphs: Extending
  high-dimensional data analysis to networks and other irregular domains.
\newblock \emph{{IEEE} Signal Process. Mag.}, 30\penalty0 (3):\penalty0 83--98,
  2013.

\bibitem[Thekumparampil et~al.(2018)Thekumparampil, Wang, Oh, and Li]{agnn}
Thekumparampil, K., Wang, C., Oh, S., and Li, L.-J.
\newblock Attention-based graph neural network for semi-supervised learning.
\newblock 03 2018.

\bibitem[Velickovic et~al.(2018)Velickovic, Cucurull, Casanova, Romero,
  Li{\`{o}}, and Bengio]{p13}
Velickovic, P., Cucurull, G., Casanova, A., Romero, A., Li{\`{o}}, P., and
  Bengio, Y.
\newblock Graph attention networks.
\newblock In \emph{ICLR}. OpenReview.net, 2018.

\bibitem[Verma et~al.(2018)Verma, Boyer, and Verbeek]{p16}
Verma, N., Boyer, E., and Verbeek, J.
\newblock Feastnet: Feature-steered graph convolutions for 3d shape analysis.
\newblock In \emph{CVPR}, pp.\  2598--2606. {IEEE} Computer Society, 2018.

\bibitem[Wu et~al.(2019{\natexlab{a}})Wu, Jr., Zhang, Fifty, Yu, and
  Weinberger]{p11}
Wu, F., Jr., A. H.~S., Zhang, T., Fifty, C., Yu, T., and Weinberger, K.~Q.
\newblock Simplifying graph convolutional networks.
\newblock In Chaudhuri, K. and Salakhutdinov, R. (eds.), \emph{ICML}, volume~97
  of \emph{Proceedings of Machine Learning Research}, pp.\  6861--6871. {PMLR},
  2019{\natexlab{a}}.

\bibitem[Wu et~al.(2019{\natexlab{b}})Wu, Tang, Zhu, Wang, Xie, and Tan]{p4}
Wu, S., Tang, Y., Zhu, Y., Wang, L., Xie, X., and Tan, T.
\newblock Session-based recommendation with graph neural networks.
\newblock In \emph{AAAI}, pp.\  346--353. {AAAI} Press, 2019{\natexlab{b}}.

\bibitem[Wu et~al.(2020)Wu, Cai, Xiao, Chen, and Ooi]{p20}
Wu, Y., Cai, S., Xiao, X., Chen, G., and Ooi, B.~C.
\newblock Privacy preserving vertical federated learning for tree-based models.
\newblock \emph{Proc. {VLDB} Endow.}, 13\penalty0 (11):\penalty0 2090--2103,
  2020.

\bibitem[Wu et~al.(2019{\natexlab{c}})Wu, Pan, Chen, Long, Zhang, and Yu]{p7}
Wu, Z., Pan, S., Chen, F., Long, G., Zhang, C., and Yu, P.~S.
\newblock A comprehensive survey on graph neural networks.
\newblock \emph{CoRR}, abs/1901.00596, 2019{\natexlab{c}}.

\bibitem[Xu et~al.(2019)Xu, Hu, Leskovec, and Jegelka]{p14}
Xu, K., Hu, W., Leskovec, J., and Jegelka, S.
\newblock How powerful are graph neural networks?
\newblock In \emph{ICLR}. OpenReview.net, 2019.

\bibitem[Yang et~al.(2019)Yang, Liu, Chen, and Tong]{p18}
Yang, Q., Liu, Y., Chen, T., and Tong, Y.
\newblock Federated machine learning: Concept and applications.
\newblock \emph{{ACM} Trans. Intell. Syst. Technol.}, 10\penalty0 (2):\penalty0
  12:1--12:19, 2019.

\bibitem[Yang et~al.(2016)Yang, Cohen, and
  Salakhutdinov]{DBLP:journals/corr/YangCS16}
Yang, Z., Cohen, W.~W., and Salakhutdinov, R.
\newblock Revisiting semi-supervised learning with graph embeddings.
\newblock \emph{CoRR}, abs/1603.08861, 2016.
\newblock URL \url{http://arxiv.org/abs/1603.08861}.

\bibitem[Yang et~al.(2020)Yang, Wang, Chen, Shi, Xu, Xu, Tian, and Xu]{p25}
Yang, Z., Wang, Y., Chen, X., Shi, B., Xu, C., Xu, C., Tian, Q., and Xu, C.
\newblock {CARS:} continuous evolution for efficient neural architecture
  search.
\newblock In \emph{CVPR}, pp.\  1826--1835. {IEEE}, 2020.

\bibitem[Yu et~al.(2018)Yu, Yin, and Zhu]{p3}
Yu, B., Yin, H., and Zhu, Z.
\newblock Spatio-temporal graph convolutional networks: {A} deep learning
  framework for traffic forecasting.
\newblock In Lang, J. (ed.), \emph{IJCAI}, pp.\  3634--3640. ijcai.org, 2018.

\bibitem[Zhao et~al.(2020)Zhao, Wei, and Yao]{p6}
Zhao, H., Wei, L., and Yao, Q.
\newblock Simplifying architecture search for graph neural network.
\newblock In Conrad, S. and Tiddi, I. (eds.), \emph{CIKM}, volume 2699 of
  \emph{{CEUR} Workshop Proceedings}. CEUR-WS.org, 2020.

\bibitem[Zhong et~al.(2018)Zhong, Yan, Wu, Shao, and Liu]{BlockQNN}
Zhong, Z., Yan, J., Wu, W., Shao, J., and Liu, C.
\newblock Practical block-wise neural network architecture generation.
\newblock In \emph{CVPR}, pp.\  2423--2432. {IEEE} Computer Society, 2018.

\end{thebibliography}
\end{document}